\documentclass[onecolumn,11pt,letterpaper]{article}  
\usepackage[english]{babel}
\usepackage{times}

\usepackage{epsfig}
\usepackage{graphicx}
\usepackage{amssymb,amsmath}
\usepackage{wrapfig}

\usepackage{subfigure}

\usepackage{algorithm}
\usepackage{algorithmic}

\usepackage{color,soul}

\graphicspath{{media/}}

\usepackage[left=1in,top=1in,right=1in,bottom=1in,nohead,foot=0.4in]{geometry}

\usepackage{url}
\usepackage[nodayofweek]{datetime}

\usepackage[sort&compress]{natbib}
\def\cite#1{\citep{#1}}

\newcommand{\captionfonts}{\it}

\makeatletter  
\long\def\@makecaption#1#2{%
  \vskip\abovecaptionskip
  \sbox\@tempboxa{{\captionfonts #1: #2}}%
  \ifdim \wd\@tempboxa >\hsize
    {\captionfonts #1: #2\par}
  \else
    \hbox to\hsize{\hfil\box\@tempboxa\hfil}%
  \fi
  \vskip\belowcaptionskip}

\renewcommand\section{\@startsection{section}{1}{\z@}%
                       {-0.5ex \@plus -0.4ex \@minus -0.4ex}%
                       {0.5ex \@plus 0.4ex \@minus 0.4ex}%
                       {\normalfont\large\bfseries\boldmath
                        \rightskip=\z@ \@plus 8em\pretolerance=10000 }}
\renewcommand\subsection{\@startsection{subsection}{2}{\z@}%
                       {-0.5ex \@plus -0.2ex \@minus -0.2ex}%
                       {0.5ex \@plus 0.2ex \@minus 0.2ex}%
                       {\normalfont\normalsize\bfseries\boldmath
                        \rightskip=\z@ \@plus 8em\pretolerance=10000 }}
\renewcommand\subsubsection{\@startsection{subsubsection}{2}{\z@}%
                       {-0.5ex \@plus -0.2ex \@minus -0.2ex}%
                       {0.5ex \@plus 0.2ex \@minus 0.2ex}%
                       {\normalfont\normalsize\bfseries\boldmath
                        \rightskip=\z@ \@plus 8em\pretolerance=10000 }}

\makeatother   

\usepackage{dsfont}
\DeclareMathOperator*{\argmin}{argmin}

\DeclareMathOperator*{\dist}{Dist}

\newcommand{\Rone}{\mathds{R}}
\newcommand{\x}{\mathbf{x}}
\newcommand{\W}{\mathbf{W}}
\begin{document}
\usdate

\title{Random Forests for Metric Learning \\with Implicit Pairwise 
Position Dependence}
\author{Caiming Xiong, David Johnson, Ran Xu and Jason J. Corso \\
Department of Computer Science and Engineering\\
SUNY at Buffalo\\
\texttt{\small \{cxiong, davidjoh, rxu2, jcorso\}@buffalo.edu}}

\date{}

\maketitle


\begin{abstract} 
Metric learning makes it plausible to learn distances for complex 
  distributions of data from labeled data.  However, to date, most 
  metric learning methods are based on a single Mahalanobis metric, 
  which cannot handle heterogeneous data well.  Those that learn 
  multiple metrics throughout the space have demonstrated superior 
  accuracy, but at the cost of computational efficiency.  Here, we 
  take a new angle to the metric learning problem and learn a single 
  metric that is able to implicitly adapt its distance function 
  throughout the feature space.  This metric adaptation is 
  accomplished by using a random forest-based classifier to underpin 
  the distance function and incorporate both absolute pairwise 
  position and standard relative position into the representation.  We 
  have implemented and tested our method against state of the art 
  global and multi-metric methods on a variety of data sets.  Overall, 
  the proposed method outperforms both types of methods in terms of 
  accuracy (consistently ranked first) and is an order of magnitude 
  faster than state of the art multi-metric methods (16$\times$ faster in the 
  worst case).
\end{abstract}

\section{Introduction}

\begin{figure*} [tb]
  \includegraphics[width=0.34\linewidth]{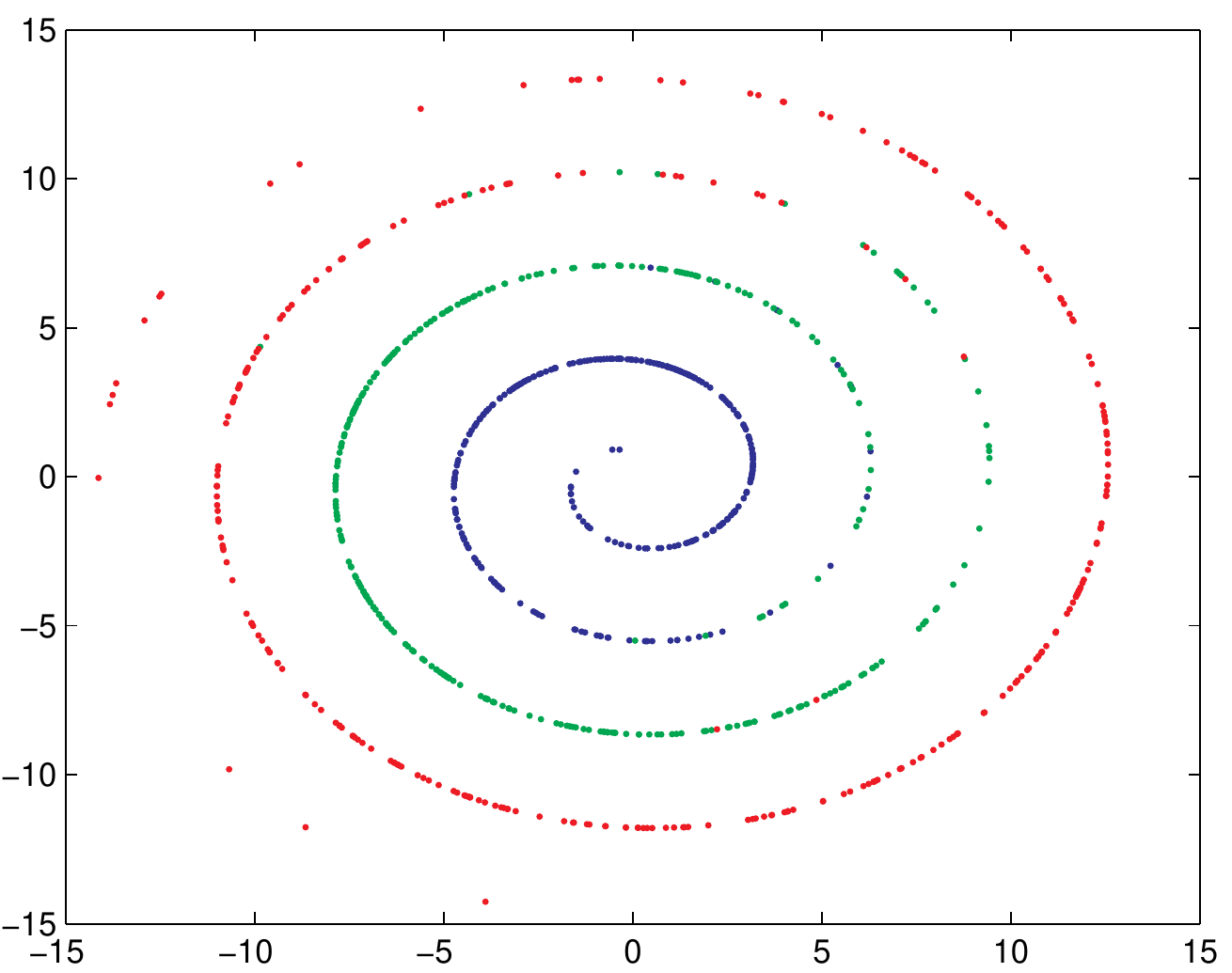}
  \includegraphics[width=0.64\linewidth]{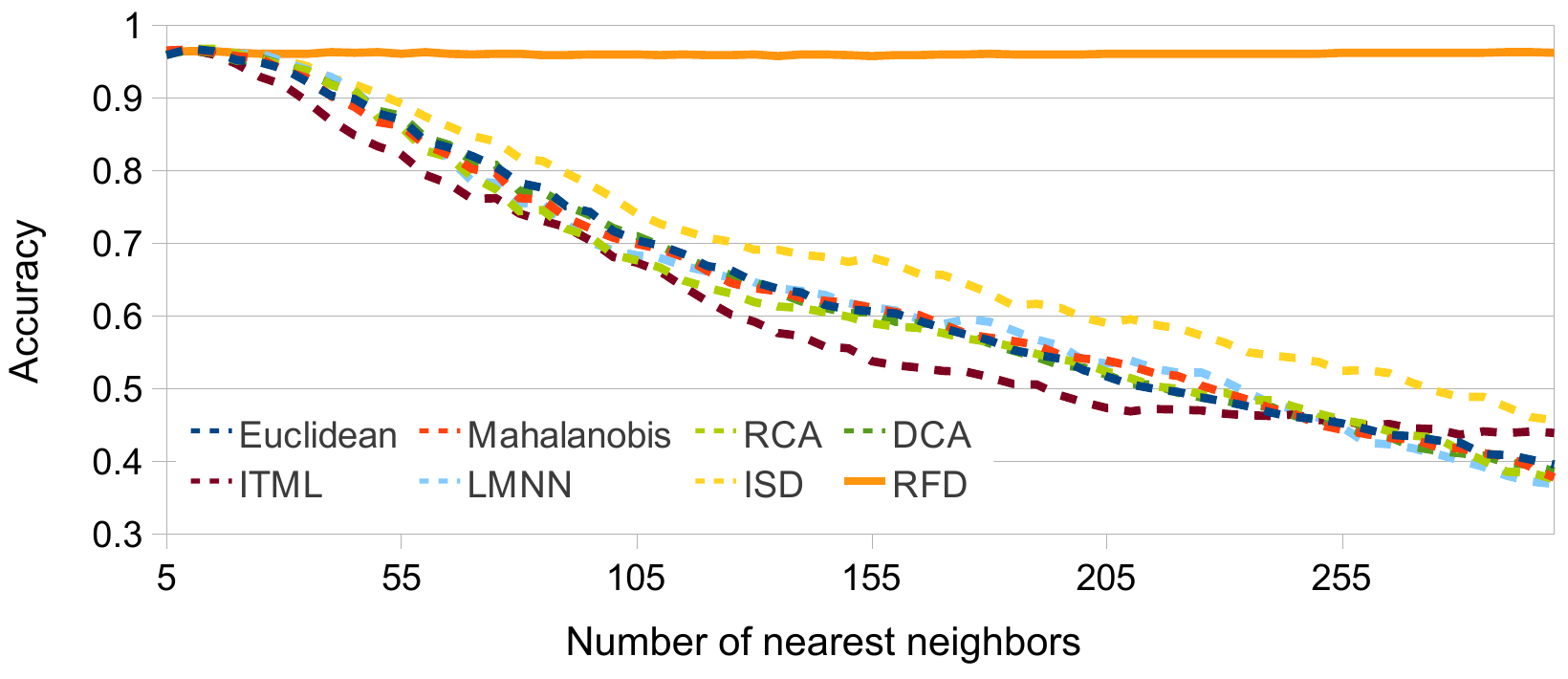}
  \caption{An example using a classic swiss roll data set comparing 
  both global and position-specific Mahalanobis-based methods with our 
  proposed method, RFD. All methods, including the baseline Euclidean, perform well at low $k$-values due to local linearity. However, as $k$ increases and the global nonlinearity of the data becomes important, the monolithic methods' inability to incorporate position information causes their performance to degrade until it is little better than chance.  The position-specific ISD method performs somewhat better, but even with a Mahalanobis matrix at every point it is unable to capture the globally nonlinear relations between points.  Our method, by comparison, shows no degradation as $k$ increases. (3 classes, 900 samples, validated using $k$-nearest neighbor classification, with varying $k$)} 
\label{fig:introcomparison}
\end{figure*}

Although the Euclidean distance is a simple and convenient metric, it is often 
not an accurate representation of the underlying shape of the data 
\cite{frome2006image}.  Such a representation is crucial in many real-world 
applications \cite{yang2011learning,boiman2008defense}, such as object 
classification \cite{fink2005object,frome2007learning1}, text document retrieval 
\cite{lebanon2006metric,wang2010text} and face verification 
\cite{chopra2005learning,nguyen2011cosine}, and methods that learn a distance 
metric from training data have hence been widely studied in recent years.
We present a new angle on the metric learning problem based on random forests 
\cite{amit1997shape, breiman2001random} as the underlying distance 
representation.  The emphasis of our work is the capability to incorporate the 
absolute position of point pairs in the input space
without requiring a separate metric per instance or exemplar.  In doing so, our 
method, called random forest distance (RFD), is able to adapt to the underlying 
shape of the data by varying the metric based on the \textit{position} of sample 
pairs in the feature space while maintaining the efficiency of a single metric.
In some sense, our method achieves a middle-ground between the two main classes 
of existing methods---single, global distance functions and multi-metric sets of 
distance functions---overcoming the limitations of both (see Figure 
\ref{fig:introcomparison} for an illustrative example).  We next elaborate upon 
these comparisons.

The metric learning literature has been dominated by methods that learn a global 
Mahalanobis metric, with representative methods 
\cite{xing2003distance,bar2003learning,hoi2006learning,davis2007information,weinberger2009distance, 
shen2010scalable,nguyen2008metric,shi2011learning}.  In brief, given a set of 
pairwise constraints (either by sampling from label data, or collecting side 
information in the semi-supervised case), indicating pairs of points that should 
or should not be grouped (i.e., have small or large distance, respectively), the 
goal is to find the appropriate linear transformation of the data to best 
satisfy these constraints.
One such method \cite{xing2003distance} minimizes the distance between 
positively-linked points subject to the constraint that negatively-linked points 
are separated, but requires solving a computationaly expensive semidefinite 
programming problem. Relevant Component Analysis (RCA) \cite{bar2003learning} 
learns a linear Mahalanobis transformation to satisfy a set of positive 
constraints.  Discriminant Component Analysis (DCA) \cite{hoi2006learning} 
extends RCA by exploring negative constraints. ITML \cite{davis2007information} 
minimizes the LogDet divergence under positive and negative linear constraints, 
and LMNN \cite{weinberger2009distance, shen2010scalable} learns a distance 
metric through the maximum margin framework. \cite{nguyen2008metric} formulate 
metric learning as a quadratic semidefinite programming problem with local 
neighborhood constraints and linear time complexity in the original feature 
space. More recently, researchers have begun developing fast algorithms that can 
work in an online manner, such as POLA \cite{shalev2004online}, MLCL 
\cite{globerson2006metric} and LEGO \cite{jain2008fast}. 

These global methods learn a single Mahalanobis metric using the relative 
position of point pairs: \\$\dist{(\x_i,\x_j)} = (\x_i-\x_j)^{\sf 
T}\W(\x_i-\x_j)$.  Although the resulting single metric is efficient, it is 
limited in its capacity to capture the shape of complex data.  In contrast, a 
second class, called multi-metric methods, distributes
distance metrics throughout the input space; in the limit, they estimate a 
distance metric per instance or exemplar, e.g., 
\cite{frome2006image,frome2007learning1} for the case of Mahalanobis metrics.  
\cite{zhan2009learning} extend \cite{frome2006image} by propagating
metrics learned on training exemplars to learn a matrix for each unlabeled point 
as well.  However, these point-based multi-metric methods all suffer from high 
time and space complexity due to the need to learn and store $\mathbf{O}(n)$ $d$ 
by $d$ metric matrices.  A more efficient approach to this second class is to 
divide the data into subsets and learn a metric for each subset 
\cite{babenko2009similarity, weinberger2008fast}.  However, these methods have 
strong assumptions in generating these subsets; for example,
\cite{babenko2009similarity} learns at most one metric per category, forfeiting 
the possibility that different samples within a category may require different 
metrics.


We propose a metric learning method that is able to achieve both the efficiency 
of the global methods and specificity of the multi-metric methods.  Our method, 
the random forest distance (RFD), transforms the metric learning problem into a 
binary classification problem and uses random forests as the underlying 
representation \cite{amit1997shape, 
breiman2001random,leistner2009semi,biau2010layered}.  In this general form, we 
are able to incorporate the position of samples implicitly into the metric and 
yet maintain a single and efficient global metric.  To that end, we use a novel 
point-pair mapping function that encodes both the position of the points 
relative to each other and their absolute position within the feature space.  
Our experimental analyses demonstrate the importance of incorporating position 
information into the metric (Section \ref{sec:experiments}).

We use the random forest as the underlying representation for several reasons.  
First, the output of the random forest algorithm is a simple ``yes'' or ``no'' 
vote from each tree in the forest.  In our case, ``no'' votes correspond to 
positively constrained training data, and ``yes'' votes correspond to negatively 
constrained training data.  The number of yes votes, then, is effectively a 
distance function, representing the relative resemblance of a point pair to 
pairs that are known to be dissimilar versus pairs that are known to be similar.
Second, random forests are efficient and scale well, and have been shown to be 
one of the most powerful and scalable supervised methods for handling 
high-dimensional data \cite{caruana2006icml}---in contrast to instance-specific 
multi-metric methods \cite{frome2006image,frome2007learning1}, the storage 
requirement of our method is independent of the size
of the input data set.  Our experimental results indicate RFD is at least 16 
times faster than the state of the art multi-metric method.  Third, because 
random forests are non-parametric, they make minimal assumptions about the shape 
and patterning of the data \cite{breiman2001random}, affording a flexible model 
that is inherently nonlinear.  In the next section, we describe the new RFD 
method in more detail, followed by a
thorough comparison to the state of the art in Section \ref{sec:experiments}.


\section{Random Forest Distance: Implicitly Position-Dependent Metric Learning}
\label{sec:main}

Our random forest-based approach is inspired by several other recent 
advances in metric learning 
\cite{shalev2004online,babenko2009similarity}  that reformulate the 
metric learning problem into a classification problem.  However, where 
these approaches restricted the form of the learned distance function 
to a Mahalanobis matrix, thus precluding the use of position 
information, we adopt a more general formulation of the classification 
problem that removes this restriction.


Given the instance set $X=\{\x_1,\x_2,\cdots,\x_N\}$, each $\x_i\in 
\Rone^m$ is a vector of $m$ features.  Taking a geometric 
interpretation of each $\x_i$, we consider $\x_i$ the 
\textit{position} of sample $i$ in the space $\Rone^m$.  The value of 
this interpretation will become clear throughout the paper as the 
learned metric will implicitly vary over $\Rone^m$, which allows it to 
adapt the learned metric based on local structure in a manner similar 
to the instance-specific multi-metric methods, e.g., \cite{frome2006image}.
Denote two pairwise constraint sets: a must-link constraint set $S 
=\{(\x_i,\x_j)| \x_i$ and $\x_j$ are similar$\}$ and a do-not-link 
constraint set $D =\{(\x_i,\x_j)|\x_i$ and $\x_j$ are dissimilar$\}$. 
For any constraint $(\x_i,\x_j)$, denote $y_{ij}$ as the ideal 
distance between $\x_i$ and $\x_j$. If $(\x_i,\x_j)\in S$, then the 
distance $y_{ij} = 0$, otherwise $y_{ij}=1$. Therefore, we seek a 
function $\dist{(\cdot,\cdot)}$ from an appropriate
function space $H$:
\begin{align}
\dist{(\cdot,\cdot)}^* =
\argmin_{\dist{(\cdot,\cdot)}\in H} 
\frac{1}{\lvert S\cup D\rvert}\sum_{(\x_i,\x_j)\in S\cup D} 
l(\dist{(\x_i,\x_j)}, y_{ij})
\enspace,
\end{align}
where $l(\cdot)$ is some loss function that will be specified by the specific 
classifier chosen.  In our random forests case, we minimize expected 
loss, as in many classification problems. So consider 
$\dist{(\cdot,\cdot)}$ to be a binary classifier for the classes $0$ 
and $1$.  For flexibility, we redefine the problem as 
$\dist{(\x_i,\x_j)}=F(\phi(\x_i,\x_j))$, where $F(\cdot)$ is some 
classification model, and $\phi(\x_i,\x_j)$ is a mapping function 
that maps the pair $(\x_i,\x_j)$ to 
a feature vector that will serve as input for the classifier function 
$F$. To train $F$, we transform each constraint pair using the mapping 
function $\{(\x_i,\x_j),y_{ij}\}\rightarrow \{\phi(\x_i,\x_j),y_{ij}\}$ and 
submit the resulting set of vectors and labels as training data.  We 
next describe the feature mapping function $\phi$.


\subsection{Mapping function for implicitly position-dependent metric learning}
\label{mappingfun}




In actuality, all metric learning methods implicitly employ a mapping 
function $\phi(\x_i,\x_j)$.  However, Mahalanobis based methods are 
restricted in terms of what features their metric solution can encode.  
These methods all learn a (positive semidefinite) metric matrix $\W$, and 
a distance function of the form $\dist{(\x_i,\x_j)} = (\x_i-\x_j)^{\sf 
T}\W(\x_i-\x_j)$, which can be reformulated as $\dist{(\x_i,\x_j)} = 
\vec[\W]^{\sf T}\vec[(\x_i-\x_j)(\x_i-\x_j)^{\sf T}]$, where 
$\vec[\cdot]$ denotes vectorization or \textit{flattening} of a 
matrix.  Mahalanobis-based methods can thus be viewed as using the 
mapping function $\phi(\x_i,\x_j) = \vec[(\x_i-\x_j)(\x_i-\x_j)^{\sf T}]$.  
This function encodes only relative position information, and the 
Mahalanobis formulation allows the use of no other features.

However, our formulation affords a more 
general mapping function:
\begin{align}
\phi(\x_i,\x_j) = 
\begin{bmatrix}
  \mathbf{u}\\
  \mathbf{v}
\end{bmatrix}
=
\begin{bmatrix}
|\x_i-\x_j|\\
\frac12 \left(\x_i+\x_j\right)
\end{bmatrix} \enspace,
\label{eq:map}
\end{align}
which considers both the relative location of the samples $\mathbf{u}$ 
as well as their absolute 
position $\mathbf{v}$.  The output feature vector is the concatenation 
of these two and in $\Rone^{2m}$.

The relative location $\mathbf{u}$ represents the same information as 
the Mahalanobis mapping function.  Note, we take the absolute value in 
$\mathbf{u}$ to enforce symmetry in the learned metric.
The primary difference between our mapping function and that of 
previous methods is thus the information contained in 
$\mathbf{v}$---the mean of the two point vectors.  It localizes each 
mapped pair to a region of the space, which allows our method to adapt 
to heterogeneous distributions of data.  It is for this reason that we 
consider our learned metric to be implicitly position-dependent.  Note the 
earlier methods that learn position-based metrics, i.e. the methods 
that learn a metric per instance such as \cite{frome2006image}, 
incorporate absolute position of each instance only, whereas we 
incorporate the absolute position of each instance pair, which adds 
additional modeling versatility.

We note that alternate encodings of the position information are 
possible but have shortcomings.  For example, we could choose to simply concatenate the position of the two points rather than average them, but this approach raises the issue of ordering the points. Using 
\(
\mathbf{v} = \begin{bmatrix}\x_i^{\sf T} & \x_j^{\sf T}\end{bmatrix}^{\sf T}
\)
would again yield a nonsymmetric feature, and an arbitrary ordering rule would not guarantee meaningful feature comparisons.
%
The usefulness of position information varies
depending on the data set.  For data that is largely linear and 
homogenous, including $\mathbf{v}$ will only add noise to the 
features, and could worsen the accuracy.  In our experiments,  
we found that for many real data sets (and particularly for more 
difficult data sets) the inclusion of $\mathbf{v}$ significantly improves the 
performance of the metric (see Section \ref{sec:experiments}).

\begin{table*}[t]
\caption{UCI data sets used for KNN-classification testing}
\begin{center}
  {\small
\scalebox{0.9}{
\begin{tabular}{|l|c|c|c||l|c|c|c|}
\hline
Dataset & Size & Dim. & No. Classes &
Dataset & Size & Dim. & No. Classes\\
\hline\hline
Balance & 625 & 4 & 3 &
Iris & 150 & 4 & 3\\
\hline
BUPA Liver Disorders & 345 & 6 & 2 &
Pima Indians Diabetes & 768 & 8 & 2\\
\hline
Breast Cancer & 699 & 10 & 2 &
Wine & 178 & 13 & 3\\
\hline
Image Segmentation & 2310 & 19 & 7 &
Sonar & 208 & 60 & 2\\
\hline
Semeion Handwritten Digits & 1593 & 256 & 10 &
\parbox{1in}{Multiple Features\\ Handwritten Digits} & 2000 & 649 & 10\\
\hline
\end{tabular}
}
 }
\label{data sets}
\end{center}
\end{table*}

\subsection{Random forests for metric learning}
\label{sec:rf}


Random forests are well studied in the machine learning literature and 
we do not describe them in any detail; the interested reader is 
directed to \cite{amit1997shape,breiman2001random}.  In brief, a 
random forest is a set of decision trees 
$\{f_t\}_{t=1}^T$ operating on a common feature space, in our case 
$\Rone^{2m}$.  To evaluate a
point-pair $(\x_i,\x_j)$, each tree independently classifies the 
sample (based on the leaf node at which the point-pair arrives) as 
similar or dissimilar (0 or 1, respectively) and the forest averages 
them, essentially regressing a distance measure on the point-pair:
\begin{align}
\dist(\x_i,\x_j) =F(\phi(\x_i,\x_j))=\frac{1}{T}\sum_{t=1}^T 
f_t(\phi(\x_i,\x_j))
 \enspace,
\end{align}
where $f_t(\cdot)$ is the classification output of tree $t$.

It has been found empirically that random forests scale well with 
increasing dimensionality, compared with other classification methods 
\cite{caruana2006icml}, and, as a decision tree-based method, they are 
inherently nonlinear. Hence, our use of them in RFD as a
regression algorithm allows for a more scalable and more flexible 
metric than is possible using Mahalanobis methods.  Moreover, the 
incorporation of position information into this classification 
function (as described in Section \ref{mappingfun}) allows the metric 
to implicitly adapt to different regions over the feature space.  In 
other words, when a decision tree in the random forest selects a node 
split based on a value of the absolute position $\mathbf{v}$ 
sub-vector (see Eq. \ref{eq:map}), then all evaluation in the sub-tree 
is \textit{localized} to a specific half-space of $\Rone^m$.  
Subsequent splits on elements of $\mathbf{v}$ further refine the 
sub-space of emphasis $\Rone^m$.  Indeed, each path through a decision 
tree in the random forest is localized to a particular (possibly 
overlapping) sub-space.  


The RFD is not technically a \textit{metric} but rather a 
\textit{pseudosemimetric}.  Although RFD can easily be shown to be 
non-negative and symmetric, it does not satisfy the triangle 
inequality (i.e., $\dist(\x_1,\x_2) \leq \dist(\x_1,\x_3) + 
\dist(\x_2,\x_3)$) or the implication that $\dist(\x_1,\x_2) = 0 \implies \x_1 = 
\x_2$, sometimes called identity of indiscernibles.  It is 
straightforward to construct examples for both of these cases.
Although this point may appear problematic, it is not uncommon in the 
metric learning literature.
For example, by necessity, no metric 
whose distance function varies across the feature space can guarantee 
the triangle inequality is satisfied.  
\cite{frome2006image,frome2007learning1} similarly cannot satisfy the 
triangle inequality.  Our method \emph{must} violate the 
triangle inequality in order to fulfill our original objective of 
producing a metric that incorporates position data.  
Moreover, our extensive experimental results 
demonstrate the capability of RFD as a distance (Section 
\ref{sec:experiments}).  




\section{Experiments and Analysis}
\label{sec:experiments}

In this section, we present a set of experiments comparing our method to state 
of the art metric learning techniques on both a range of UCI data sets (Table 
\ref{data sets}) and an image data set taken from the Corel database.  To 
substantiate our claim of computational efficiency, we also provide an analysis 
of running time efficiency relative to an existing position-dependent metric 
learning method. 

For the UCI data sets, we compare performance at the $k$-nearest 
neighbor classification task against both standard Mahalanobis methods 
and point-based position-dependent methods.  For the former, we test 
$k$-NN classification accuracy at a range of k-values (as in Figure 
\ref{fig:introcomparison}), while the latter relies on results 
published by other methods' authors, and thus uses a fixed $k$.
For the image data set, we measure accuracy at $k$-NN retrieval, rather 
than $k$-NN classification.  We compare our results to several 
Mahalanobis methods.

The following is an overview of the primary experimental findings to 
be covered in the following sections.

\begin{figure*}[p]
\begin{center}
  \includegraphics[width=0.8\linewidth]{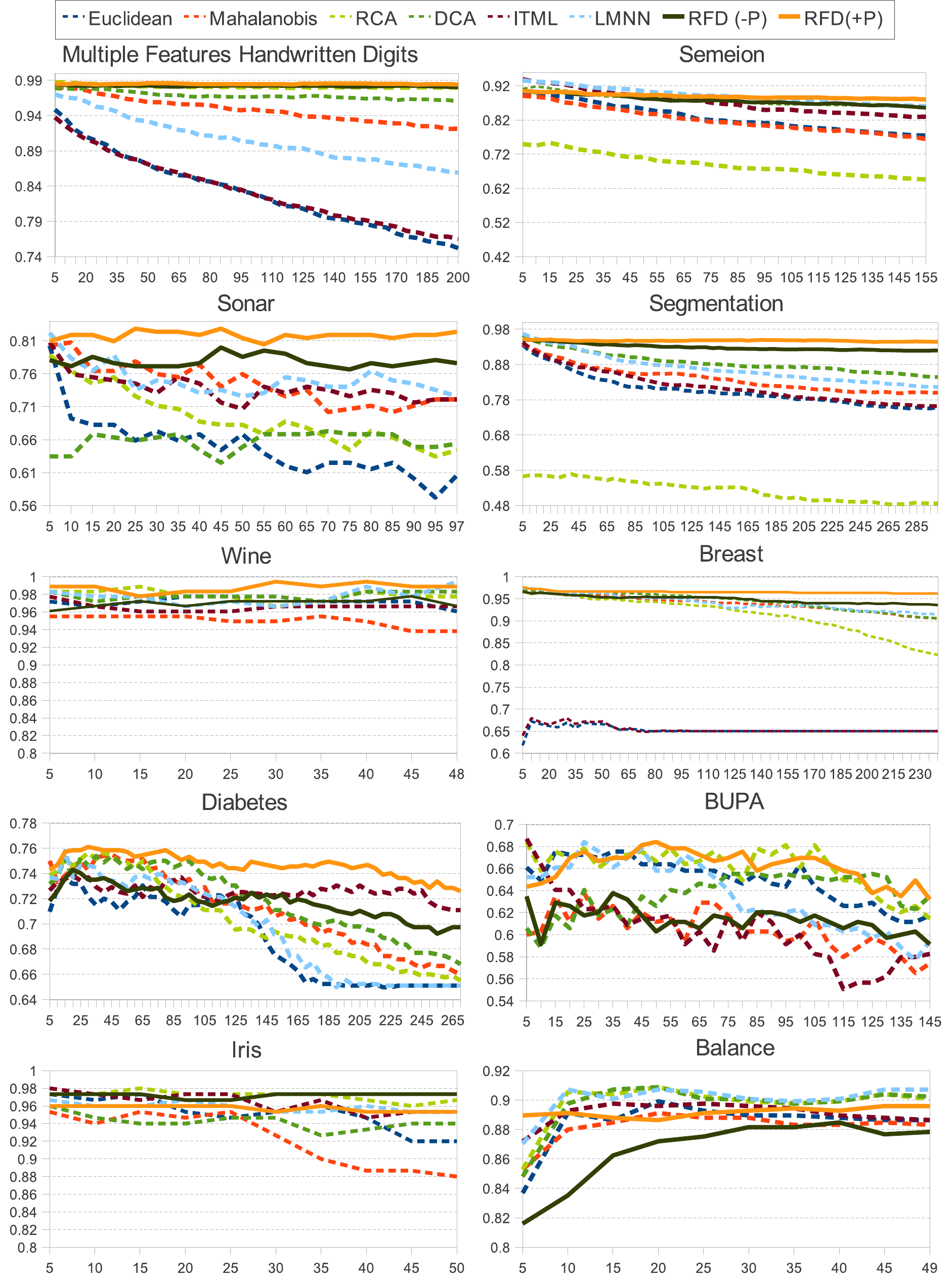}
  \caption{$k$-nearest neighbor classification results with varying 
  $k$ values of RFD versus 
  assorted global Mahalanobis methods on 10 UCI data sets .  Plots show $k$-nearest neighbor $k$-value versus accuracy.  Note in particular the segmentation and breast datasets, where RFD shows little or no degradation over increasing distances, while other methods steadily decline in accuracy.  Also note that the inclusion of position information in the RFD yields higher performance on all but the low-dimensional and highly linear iris dataset.
  \label{fig:UCIresults}
  \label{fig:uci1}
  }
  \end{center}
\end{figure*}

\begin{enumerate}
  \item RFD has the best overall performance on ten UCI data sets 
    ranging from 4 to 649 dimensions against four state of 
    the art and two baseline global Mahalanobis-based methods (Figure \ref{fig:uci1} and 
    Table \ref{tbl:ucirank}).

  \item RFD has comparable or superior accuracy to state of the art position-specific 
    methods (Table \ref{tbl:isd}).

  \item RFD is 16 to 85 times faster than the state of the art 
    position-specific method (Table \ref{tbl:time}).

  \item RFD outperforms the state of the art in nine out of ten 
    categories in the benchmark Corel image 
    retrieval problem (Figure \ref{fig:corel}).

\end{enumerate}

\subsection{Comparison with global Mahalanobis metric learning methods}
\label{sec:expglobal}

We first compare our method to a set of state of the art Mahalanobis 
metric learning methods: RCA \cite{bar2003learning}, 
DCA \cite{hoi2006learning}, Information-Theoretic Metric Learning 
(ITML)\cite{davis2007information} and distance metric learning for 
large-margin nearest neighbor classification 
(LMNN) \cite{weinberger2009distance, shen2010scalable}.  For our 
method, we test using the full feature mapping including relative 
position data, $\mathbf{u}$, and absolute 
pairwise position 
data, $\mathbf{v}$, (RFD ($+$P)) as well as with only relative position data, 
$\mathbf{u}$, (RFD ($-$P)).  To 
provide a baseline, we also show results using both the Euclidean 
distance and a heuristic Mahalanobis metric, where the $\W$ used is 
simply the covariance matrix for the data. All algorithm code was 
obtained from authors' websites, for which we are indebted (\textit{our code is 
available on \url{http://www.cse.buffalo.edu/~jcorso}}).

We test each algorithm on a number of standard small to medium scale UCI data 
sets (see Table \ref{data sets}).  All algorithms are trained using 1000 
positive and 1000 negative constraints per class, with the exceptions of RCA, 
which used only the 1000 positive constraints and LMNN, which used the full 
label set to actively select a (generally much larger) set of constraints; 
constraints are all selected randomly according to a uniform distribution.  In 
each case, we set the number of trees used by our method to 400 (\textit{see 
Section \ref{forestsize} for a discussion of the effect of varying forest 
sizes}).

Testing is performed using 5-fold cross validation on the $k$ nearest-neighbor 
classification task.  Rather than selecting a single $k$-value for this task, we 
test with varying $k$s, increasing in increments of 5 up to the maximum possible 
value for each data set (i.e. the number of elements in the smallest class).
By varying $k$ in this way, we are able to gain some insight into each method's 
ability to capture the global variation in a data set.
When $k$ is small, most of the identified neighbors lie within a small local 
region surrounding the query point, enabling linear metrics to perform fairly 
well even on globally nonlinear data by taking advantage of local linearity.  
However, as $k$ increases, local linearity becomes less practical, and the 
quality of the metric's representation of the global structure of the data is 
exposed.  Though the accuracy results at higher $k$ values do not have strong 
implications for each method's efficacy for the specific task of $k$-NN 
classification (where an ideal $k$ value can just be selected by 
cross-validation), they do indicate overall metric performance, and are highly 
relevant to other tasks, such as retrieval.

Figure \ref{fig:uci1} show the accuracy plots for ten UCI datasets.  RFD is consistently near 
the top performers on these various data sets.  In the lower dimension 
case (Iris), most methods perform well, and RFD without position 
information outperforms RFD with position information (this is the 
sole data set in which this occurs), which we attribute to the limited 
data set size (150 samples) and the position information acting as a 
distractor in this small and highly linear case.  In all other cases, the 
RFD with absolute position information significantly outperforms RFD 
without it.  In many of the more more difficult cases
(Diabetes, Segmentation, Sonar), RFD with position information significantly 
outperforms the field.   This result is suggestive that RFD can scale  
well
with increasing dimensionality, which is consistent with the findings from 
the literature that random forests are one of the most robust 
classification methods for high-dimensional data 
\cite{caruana2006icml}.

Table \ref{tbl:ucirank} provides a summary 
statistic of the methods by computing the mean-rank (lower better) 
over the ten data sets at varying $k$-values.  For all but one value of 
$k$, RFD with absolute position information has the best mean rank of 
all the methods (and for the off-case, it is ranked a close second).  
RFD without absolute position information performs comparatively 
poorer, underscoring the utility of the absolute position information.  
In summary, the results in Table \ref{tbl:ucirank} show that RFD is 
consistently able to outperform the state of the art in global metric 
learning methods on various benchmark problems.

\begin{figure*}[th!]
\begin{center}
\includegraphics[width=0.95\linewidth]{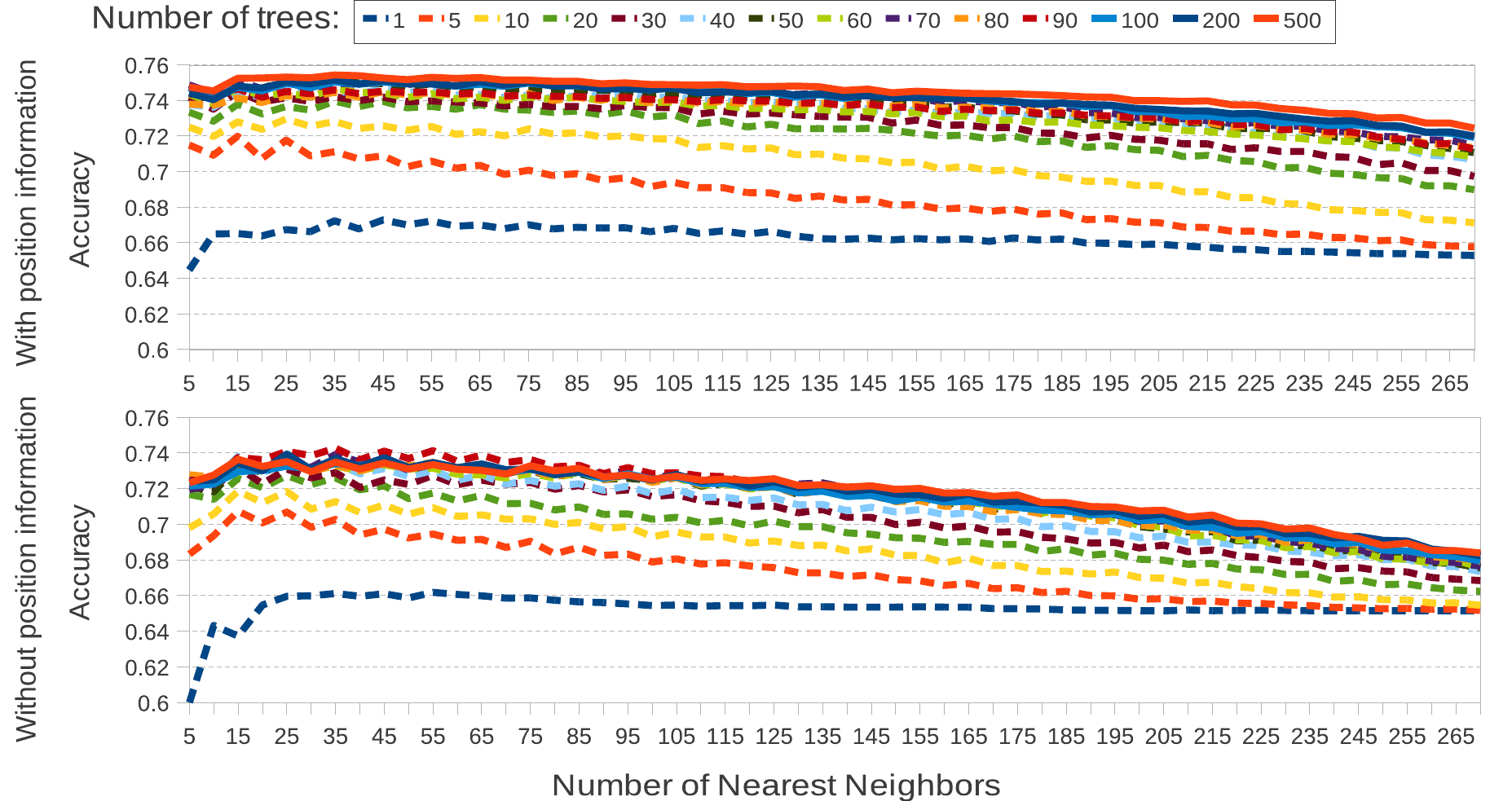}
\caption{Effect of forest size on RFD performance on the UCI diabetes data set.  Results were obtained by averaging results from 10 runs, each using 5-fold cross validation. Both with and without position information, increasing forest size yields notable improvements in accuracy up to about 100 trees.  If no position information is included, then additional trees beyond this point provide modest gains at best.  With position information, larger forests do appear to allow more fine-tuning, and can produce noticable improvements up to at least 500 trees. \label{fig:forestsize}}
\end{center}
\end{figure*}

\begin{table*}[tbp]
\caption{Mean $k$-nearest neighbor classification accuracy ranking on 
10 UCI data sets at varying $k$ values (lower rank is better). The 
mean ranking is shown in each 
table cell as well as its rank, in parentheses; i.e., for $k$ of 5, 
RFD (+P) has a mean rank of $2.9$, the number 1 mean rank.   
As expected Euclidean always has the worst rank. 
RFD with absolute position information attains the best rank in nearly all
cases, and the relative performance of both RFD methods improves as $k$ increases.}
\begin{center}
  {\small
\begin{tabular}{|l||c|c|c|c|c|c|c|c|}
\hline
         $k$-value &  Euclid & Mahal &        RCA &        DCA &       
         ITML &       LMNN &  RFD ($-$P) &   RFD ($+$P) \\
\hline
\hline
         5 &        5.8 (8) &        5.7 (7) &        4.3 (4) &        4.8 (5) &        3.9 (3) &        3.2 (2) &        5.4 (6) &       \bf{2.9 (1)} \\
\hline
        10 &        6.1 (8) &        5.6 (7) &        3.7 (3) &        4.6 (4) &        4.8 (5) &        \bf{2.9 (1)} &        5.1 (6) &        3.2 (2) \\
\hline
        15 &        5.7 (8) &        5.4 (6) &        3.9 (3) &        4.7 (5) &        5.6 (7) &        3.1 (2) &        4.6 (4) &          \bf{3 (1)} \\
\hline
        20 &        5.6 (8) &        5.4 (7) &        3.8 (3) &        5.2 (5) &        5.3 (6) &        3.7 (2) &        4.5 (4) &        \bf{2.5 (1)} \\
\hline
        25 &        6.1 (8) &        5.3 (6) &          4 (3) &        4.5 (4) &        5.4 (7) &        3.4 (2) &        4.8 (5) &        \bf{2.5 (1)} \\
\hline
        30 &        5.8 (7) &        5.9 (8) &        4.5 (5) &        4.3 (3) &        5.3 (6) &        3.5 (2) &        4.3 (3) &        \bf{2.4 (1)} \\
\hline
        35 &        5.8 (8) &        5.4 (6) &        4.3 (4) &        4.9 (5) &        5.5 (7) &          4 (3) &        3.8 (2) &        \bf{2.3 (1)} \\
\hline
        45 &        6.6 (8) &        5.5 (6) &        4.4 (4) &        4.4 (4) &        5.9 (7) &        3.3 (2) &        4.1 (3) &        \bf{1.8 (1)} \\
\hline
       Max &        6.5 (8) &        6.1 (7) &        5.1 (5) &        3.7 (3) &        5.5 (6) &        3.7 (3) &        3.5 (2) &        \bf{1.9 (1)} \\
\hline
\end{tabular}
\label{tbl:ucirank}
}
\end{center}
\end{table*}

\subsection{Varying forest size}
\label{forestsize}

One question that must be addressed when using RFD is how many trees must or should be learned in order to obtain good results.  Increasing the size of the forest 
increases computation and space requirements, and past a certain point 
yields little or no improvement and may possibly over-train.  It is 
beyond the scope of this paper to provide a full answer as to how many 
trees are needed in RFD, but we have made some observations.

First, the addition of absolute position information noticeably 
increases the benefit that may be obtained from additional trees (see 
Figure \ref{fig:forestsize}).  This result is unsurprising, 
considering the increased size of the feature vector, as well as the 
increased degree of fine-tuning possible for a metric that can vary 
from region to region.  Second, in our experiments we observe 
significant improvements in accuracy up to about 100 trees, even 
without position information, and would recommend this as a reasonable 
minimum value.  It seems reasonable that larger constraint-sets will 
require larger forests, and similarly, the more complex the shape of 
the data, the larger the forest may need to be.  But, these two points 
have not yet been thoroughly explored by our group.

\begin{table*}[p]
\caption{Comparison of test error (mean + STD) for position-dependent metric learning methods. The best performance on each data set is shown in bold.  We note that our method yields the best accuracy on 3 out of 5 data sets tested, and is within 1\% of the best on the remaining 2.}
\begin{center}
  {\small
\scalebox{1.0}{
\begin{tabular}{|l||c|c|c|c|c|c|c|c|}
\hline
Dataset & RFD & ISD L1 & ISD L2 & FSM   & FSSM  \\ 
\hline\hline
Balance &.120$\pm$.024  &\bf{.114$\pm$.013}	&.116$\pm$.014 &0.134$\pm$.020 &0.143$\pm$.013\\
Diabetes & \bf{.241$\pm$.028}  &.287$\pm$.019 &	.269$\pm$.023 &.342$\pm$.050 &.322$\pm$.232\\
Breast(Scaled) &\bf{.030$\pm$.011}&.0.31$\pm$.010 &\bf{.030$\pm$.010}	&.102$\pm$.041	&.112$\pm$.029\\
German&.277$\pm$.039	&.277$\pm$.015	&\bf{.274$\pm$.013}	&.275$\pm$.021	&0.275$\pm$.060\\
Haberman&\bf{.273$\pm$.029}	&.277$\pm$.029	&\bf{.273$\pm$.025}	&.276$\pm$.032	&.276$\pm$.029\\
\hline
\end{tabular}
}
\label{tbl:isd}
}
\end{center}
\end{table*}

\begin{table*}[p]
\caption{Run-time comparison of ISD and RFD (with position 
information, using 1000 trees) across several UCI data sets.  All times 
are in seconds.  Results were obtained by performing 5-fold cross 
validation and averaging the time for each fold. $^*$Note that ISD is multithreading across 12 cores, while our implementation of RFD is fully sequential.\label{tab:isdtime}}
\begin{center}
  {\small
\begin{tabular}{|l||c|c|c|}
\hline
Dataset & ISD Time$^*$ & RFD Time & ISD:RFD Ratio\\
\hline\hline
Iris & 34.6 &	2.1 &	16.4\\
Balance & 620.3 & 11.2 &	55.3\\
Breast (scaled) &	657.4 &	7.8 &	84.6\\
Diabetes & 849.5 &	14.7 &	57.8\\
\hline
\end{tabular}
\label{tbl:time}
}
\end{center}
\end{table*}

\subsection{Comparison with position-specific multi-metric methods}

We compare our method to three multi-metric methods that 
incorporate absolute position (via instance-specific metrics): 
FSM, FSSM and ISD.  FSM \cite{frome2006image} learns an 
instance-specific distance for each labeled example. FSSM 
\cite{frome2007learning1} is an extension of FSM that enforces global 
consistency and comparability among the different instance-specific 
metrics. ISD \cite{zhan2009learning} first learns instance-specific 
distance metrics for each labeled data point, then uses metric 
propagation to generate instance-specific metrics for unlabeled points 
as well.

We again use the ten UCI data sets, but under the same conditions used 
by these methods' authors.  Accuracy is measured on the $k$-NN
task ($k$=11) with three-fold cross validation. The parameters of 
the compared methods are set as suggested in \cite{zhan2009learning}. 
Our RFD method chooses 1\% of the available positive constraints and 
1\% of the available negative constraints, and constructs a random 
forest with 1000 trees. We report the average result of ten different 
runs on each data set, with random partitions of training/testing data 
generated each time (see Table \ref{tbl:isd}).
These results show that our RFD method yields performance better than 
or comparable to state of the art explicitly multi-metric 
learning methods. Additionally, because we only learn one distance 
function and random forests are an inherently efficient technique, our 
method offers significantly better computational efficiency than these 
instance-specific approaches (see Table \ref{tbl:time})---between 16 
to 85 times faster than ISD.

The comparable level of accuracy is not surprising.  While our method 
is a single metric in form, in practice its implicit 
position-dependence allows it to act like a multi-metric system.  
Notably, because our method learns using the position of each 
point-pair rather than each point, it can potentially encode up to 
$n^2$ implicit position-specific metrics, rather than the 
$\mathbf{O}(n)$ learned by existing position-dependent methods, which 
learn a single metric per instance/position.  RFD 
is a stronger way to learn a position-dependent metric, because even 
explicit multi-metric methods will fail over global distances in 
cases where a single (Mahalanobis) metric cannot capture the 
relationship between its associated point and every other point in the 
data.

\begin{figure*}[t!]
\begin{center}
   \includegraphics[width=0.95\linewidth]{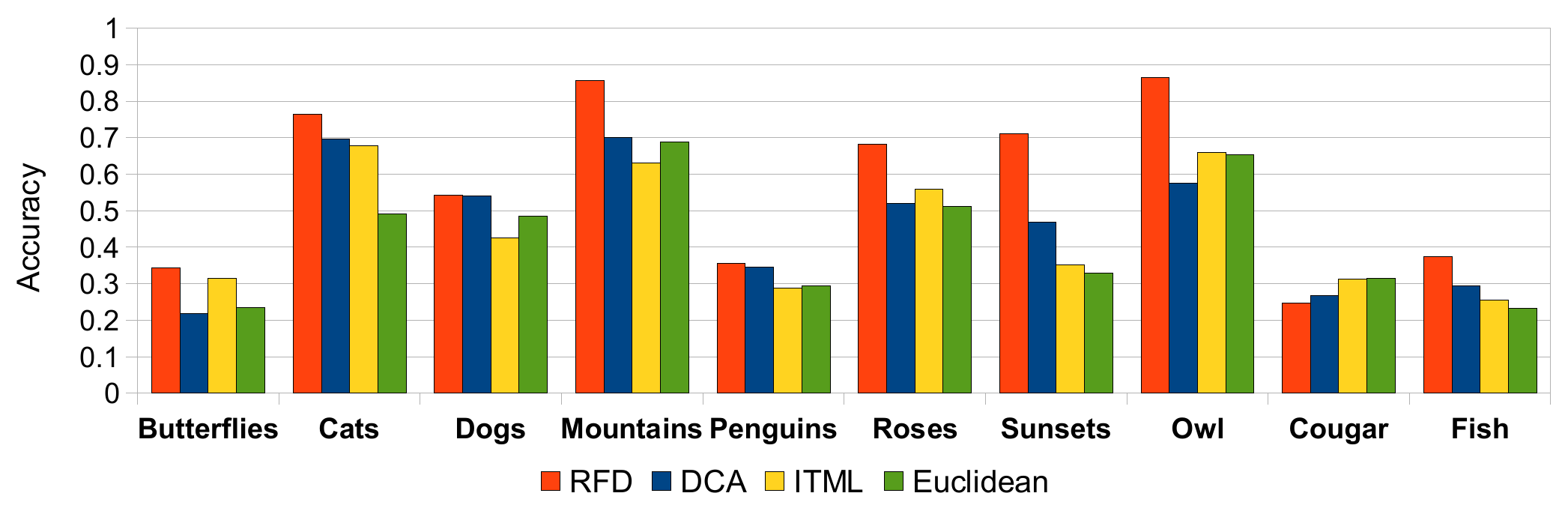}
   \caption{Average retrieval precision on top 20 nearest neighbors of 
   images in the Corel data set.  RFD outperforms DCA, ITML and the 
   baseline Euclidean measure on all but one 
   category.\label{fig:corel}}
\end{center}
\end{figure*}

\subsection{Retrieval on the Corel image data set}

We also evaluate our method's performance on the challenging image 
retrieval task because this task differs from $k$-NN classification by 
emphasizing the accuracy of individual pairwise distances rather than 
broad patterns.  For this task, we use an image data set taken from the 
Corel image database.  We select ten image categories of varying types 
(cats, roses, mountains, etc.---the classes and images are similar to 
those used by Hoi et al. to validate DCA \cite{hoi2006learning}), each 
with a clear semantic meaning.  Each class contains 100 images, for a 
total of 1000 images in the data set.

For each image, we extract a 36-dimensional low-level feature vector 
comprising color, shape and texture. For color, we extract mean, 
variance and skewness in each HSV color channel, and thus obtain 9 
color features. For shape, we employ a Canny edge detector and 
construct an 18-dimensional edge direction histogram for the image.  
For  texture, we apply Discrete Wavelet Transformation (DWT) to 
graylevel versions of original RGB images. A Daubechies-4 wavelet 
filter is applied to perform 3-level decomposition, and mean, variance 
and mode of each of the 3 levels are extracted as a 9-dimensional 
texture feature.

We compare three state of the art algorithms and a Euclidean distance 
baseline: ITML, DCA, and our RFD method (with absolute position 
information).  For ITML, we vary the parameter $\gamma$ from $10^{-4}$ 
to $10^{4}$ and choose the best ($10^{-3}$).  For each method, we generate 1\% 
of the available positive constraints and 1\% of the available 
negative constraints (as proposed in \cite{hoi2006learning}). For RFD, we construct a random forest 
with 1500 trees.
Using five-fold cross validation, we retrieve the 20 nearest neighbors 
of each image under each metric.  Accuracy is determined by counting 
the fraction of the retrieved images that are the same class as the 
image that retrieved them. 
We repeat this experiment 10 times with differing random folds and report the average results in Figure \ref{fig:corel}.  RFD clearly outperforms the other methods tested, achieving the best accuracy on all but the cougar category.  Also note that ITML performs roughly on par with or worse than the baseline on 7 classes, and DCA on 5, while RFD fails only on 1, indicating again that RFD provides a better global distance measure than current state of the art approaches, and is less likely to sacrifice performance in one region in order to gain it in another.

\section{Conclusion}
\label{sec:conclusion}

In this paper, we have proposed a new angle to the metric learning problem.  Our 
method, called random forest distance (RFD), incorporates both conventional 
relative position of point pairs as well as absolute position of point pairs 
into the learned metric, and hence implicitly adapts the metric through the 
feature space.  Our evaluation has demonstrated the capability of RFD, which has 
best overall performance in terms of accuracy and speed on a variety of 
benchmarks.



There are immediate directions of inquiry that have been paved with this paper.  
First, RFD further demonstrates the capability of classification methods 
underpinning metric learning.  Similar feature mapping functions and other 
underlying forms for the distance function need to be investigated.
Second, the utility of absolute pairwise position is clear from our work, which 
is a good indication of the need for multiple metrics.  Open questions remain 
about other representations of the position as well as the use of position in 
other metric forms, even the classic Mahalanobis metric.  Third, there are 
connections between random forests and nearest-neighbor methods, which may 
explain the good performance we have observed.  We have not explored them in any 
detail in this paper and plan to in the future.  Finally, we are also 
investigating the use of RFD on larger-scale, more diverse data sets like the 
new MIT SUN image classification data set.








\section*{Acknowledgements}

We are grateful for the support in part provided through the following grants: NSF CAREER IIS-0845282, ARO YIP W911NF-11-1-0090, DARPA Mind’s Eye W911NF-10-2-0062, DARPA CSSG D11AP00245, and NPS N00244-11-1-0022.  Findings are those of the authors and do not reflect the views of the funding agencies.

{

}

\end{document}